\documentclass[aps,prx,reprint,preprintnumbers,superscriptaddress,nofootinbib,longbibliography,floatfix]{revtex4-2}
\pdfoutput=1
\usepackage{rotating}
\usepackage{array}
\usepackage{amsmath}
\usepackage[normalem]{ulem}
\usepackage{slashed}
\usepackage{booktabs}
\usepackage[pdftex,table]{xcolor}
\usepackage{xfrac}
\usepackage{mathtools}
\usepackage{empheq}
\usepackage{multirow}
\usepackage{amssymb}
\usepackage{url}
\usepackage{comment}
\usepackage[caption=false]{subfig}
\usepackage[normalem]{ulem}

\usepackage{hyperref}
\hypersetup{
  colorlinks=true,
  citecolor=blue,
  linkcolor=blue,
  urlcolor=blue
}

\begin{document}

\title{Comparison of Point Cloud and Image-based Models for Calorimeter Fast Simulation}

\author{Fernando Torales Acosta}
\email{ftoralesacosta@lbl.gov}
\affiliation{Physics Division, Lawrence Berkeley National Laboratory, Berkeley, CA 94720, USA}

\author{Vinicius Mikuni}
\affiliation{National Energy Research Scientific Computing Center, Berkeley Lab, Berkeley, CA 94720, USA}

\author{Benjamin Nachman}
\affiliation{Physics Division, Lawrence Berkeley National Laboratory, Berkeley, CA 94720, USA}
\affiliation{Berkeley Institute for Data Science, University of California, Berkeley, CA 94720, USA}

\author{Miguel Arratia}
\affiliation{Department of Physics and Astronomy, University of California, Riverside, CA 92521, USA}
\affiliation{Thomas Jefferson National Accelerator Facility, Newport News, Virginia 23606, USA}

\author{Bishnu Karki}
\affiliation{Department of Physics and Astronomy, University of California, Riverside, CA 92521, USA}

\author{Ryan Milton}
\affiliation{Department of Physics and Astronomy, University of California, Riverside, CA 92521, USA}

\author{Piyush Karande}
\affiliation{Computational Engineering Division, Lawrence Livermore National Laboratory, Livermore CA 94550}

\author{Aaron Angerami}
\affiliation{Nuclear and Chemical Science Division, Lawrence Livermore National Laboratory, Livermore, CA 94550}

\begin{abstract}
    Score based generative models are a new class of generative models that have been shown to accurately generate high dimensional calorimeter datasets. Recent advances in generative models have used images with 3D voxels to represent and model complex calorimeter showers.  Point clouds, however, are likely a more natural representation of calorimeter showers, particularly in calorimeters with high granularity. Point clouds preserve all of the information of the original simulation, more naturally deal with sparse datasets, and can be implemented with more compact models and data files. In this work, two state-of-the-art score based models are trained on the same set of calorimeter simulation and directly compared.

\end{abstract}

\maketitle


\tableofcontents

\section{Introduction}
\label{sec:intro}

Detector simulations are essential tools for data analysis by connecting particle and nuclear physics predictions to measurable quantities.  The most precise detector simulations are computationally expensive.  This is especially true for calorimeters, which are designed to stop most particles and thus require modeling interactions from the highest accessible energies down to the lowest ones.  Well-established experiments typically have bespoke fast simulations that capture the salient aspects of the precise simulations (usually based on \textsc{Geant}~\cite{Geant4,Geant4-add1,Geant4-add2}) at a fraction of the computational cost.  Traditionally, fast simulations are constructed to reproduce a series of low-dimensional observables.  Furthermore, assembling an effective fast simulation is time intensive.  If there was a way to build a fast simulation automatically and using the full detector dimensionality, then data analysis at existing and developing experiments could be greatly enhanced.

Deep learning (DL) has been used to build automated and high-dimensional fast simulations (`surrogate models') for calorimeters.  Starting from Generative Adversarial Networks (GANs)~\cite{GANs}~\cite{GanPhys2,GanPhys3,deOliveira:2017rwa,Erdmann:2018kuh,Erdmann:2018jxd,Belayneh:2019vyx,Vallecorsa:2019ked,SHiP:2019gcl,Chekalina:2018hxi,Carminati:2018khv,Vallecorsa:2018zco,Musella:2018rdi,Deja:2019vcv,ATLAS:2022jhk,ATL-SOFT-PUB-2018-001,ATLAS:2021pzo} and now including Variational Autoencoders~\cite{VAEs}~\cite{ATL-SOFT-PUB-2018-001,ATLAS:2022jhk,Buhmann:2021lxj,Buhmann:2021caf,Diefenbacher:2023prl}, Normalizing Flows~\cite{NFs}~\cite{caloflow1,caloflow2,Buckley:2023rez,Krause:2022jna,Diefenbacher:2023vsw,Cresswell:2022tof, Liu:2023lnn}, and Diffusion Models~\cite{scoremodels}~\cite{mikuni:caloscore,Buhmann:2023bwk}, deep learning based calorimeter simulations have rapidly improved over the last years.  They are even starting to be used in actual experimental workflows, such as the ATLAS Collaboration fast simulation~\cite{ATLAS:2021pzo}.  The recent CaloChallenge~\cite{calochallenge} community comparison showcased the state-of-the-art methods deployed to increasingly granular current and future detectors.  As segmented detectors, calorimeters are naturally represented as (possibly irregular) images.  Nearly all proposed methods for DL-based calorimeter simulations are based on an image format (fixed grid of pixels).  However, these data are unlike natural images in a number of ways, most notably in their sparsity.  As such, image-based approaches pioneered in industry may not be the most effective for particle interactions.

Since most cells in a calorimeter image are empty, a more natural representation of these data may be a point cloud.  Point clouds are a set of attributes assigned to locations in space.  In the calorimeter case, the attribute is energy and the location is the cell coordinates. A calorimeter point cloud would require far fewer numbers to specify than an image representation, since only cells with non-zero energy would be recorded. The main challenges for point cloud models in contrast to image-based approaches is that they must cope with variable-length outputs that respect permutation invariance.  With a lag compared to image-based approaches, point cloud generative models for particle/nuclear physics applications have seen a rapid development in recent years~\cite{Kansal:2021cqp,Buhmann:2023pmh,Kach:2022qnf,Verheyen:2022tov,mikuni:point_clouds,Leigh:2023toe}.  However, until recently, these models have never been applied to calorimeter simulations.  

The first (and until now, only) publication describing point cloud generative models applied to calorimeters is Ref.~\cite{Buhmann:2023bwk}, which proposed generating \textsc{Geant} `hits' (deposits of energy) prior to their discritization into cells.  This innovative idea enables the separation of material interactions from readout geometry.  However, the number of hits vastly exceeds the number of non-zero cells which makes this task difficult.  In this paper, we explore point cloud generative models applied directly to cell-level information.  In other words, we take calorimeter images and compare state-of-the-art generative models that represent the same inputs as either images or (zero-suppressed) point clouds. As a case study, the two representations are compared using simulations of a high-granularity hadronic calorimeter, similar to the design planned for the ePIC detector at the future Electron-Ion Collider~\cite{EICYR, Bock:2022lwp, Insert}.

This paper is organized as follows.  Section~\ref{sec:DLmodels} describes the DL models used for the comparison.  Both the image-based and point-cloud representations are generated with diffusion models in order to make the comparison as direct as possible.  The simulation of the calorimeter dataset is found in Sec.~\ref{sec:data}. Discussion of the advantages and disadvantages of both representation, as well as numerical results are presented in Sec.~\ref{sec:results}.  The paper ends with conclusions and outlook in Sec.~\ref{sec:conclusions}.

\section{Deep Learning Models}
\label{sec:DLmodels}

Generative models for detector simulation aim to precisely emulate physics-based models, like those based on \textsc{Geant}, but using far less time than the full simulation. With $\mathcal{O}$(100) detector components, neural network architectures solely based on fully connected layers can efficiently produce high fidelity samples, resulting in surrogate models thousands of times faster than the standard simulation routines~\cite{caloflow2,ATLAS:2022jhk,ATL-SOFT-PUB-2018-001,ATLAS:2021pzo}. For higher detector granularity ($\mathcal{O}$(1k) - $\mathcal{O}$(10k)), the use of data symmetries becomes crucial to achieve precision. These can be directly included in the model design through dedicated neural network architectures or included in the data pre-processing~\cite{caloflow1}. For generative models such as normalizing flows, introducing flexible network architectures is often not trivial as the model invertibility and tractable Jacobian of the transformation places a strong restriction on the model design. A second difficulty is to achieve a stable training routine of the surrogate model. At finer granularities, neural network models tend to become larger to accommodate the data complexity, often resulting in unstable training schedules. This issue becomes more prominent in generative models such as variational autoencoders, where the latent space can vary rapidly, leading to an unstable response of the decoder network, or GANs, where the adversarial training requires careful tuning of the model hyperparameters to achieve a stable training.

Diffusion models are a class of generative neural networks that allow for stable training paired with high flexibility in the model design. Data is slowly perturbed over time using a time parameter $t \in \mathbb{R}$ that determines the perturbation level. The task of the neural network is to approximate the gradients of the log probability of the data, or the score function $\nabla_\textbf{x}p(\textbf{x}) \in \mathbb{R}^D$, based on data observations $\textbf{x} \in \mathbb{R}^D$ in the $D$-dimensional space. This can be approximated by a denoising score-matching strategy~\cite{score_denoising}. In the implementation used in this paper, data observations $\textbf{x}\sim p_{\mathrm{data}}(\textbf{x})$ are perturbed using the kernel $\mathbf{x}_t\sim q(\mathbf{x}_t|\mathbf{x})=\mathcal{N}(\mathbf{x}_t;\alpha_t\mathbf{x},\sigma_t^2\mathbf{I})$, with time-dependent parameters $\alpha$ and $\sigma$ determining the strength of the perturbation to be applied. In the variance-preserving setting of diffusion processes, $\sigma_t^2 = 1 - \alpha_t^2$. For the time-dependence, a cosine schedule is used such that $\alpha_t = \cos(0.5\pi t)$. 
The loss function to be minimized is implemented using a \textit{velocity} parameterization:
\begin{equation}
    \mathcal{L}_\theta = \mathbb{E}_{\mathbf{\epsilon},t} \left\| \mathbf{v}_t - \hat{\mathbf{v}}_{t,\theta}\right\|^2,
\end{equation}
where the time-dependent network output with trainable parameters $\theta$, $\hat{\mathbf{v}}_{t,\theta}$, is compared with the velocity of the perturbed data at time $t$, $\mathbf{v}_t \equiv \alpha_t\mathbf{\epsilon}-\sigma_t\mathbf{x}$, with $\mathbf{\epsilon}\sim \mathcal{N}(\mathbf{0},\mathbf{I})$. The score function is then identified as
\begin{equation}
    \nabla_x\log \hat{p}_\theta(\mathbf{x}_t) =  -\mathbf{x}_t - \frac{\alpha_t}{\sigma_t}\hat{\mathbf{v}}_{t,\theta}(\mathbf{x}_t).
\end{equation}

The data generation from the trained diffusion models is implemented using the DDIM sampler proposed in Ref.~\cite{DBLP:journals/corr/abs-2010-02502} that can be interpreted as an integration rule~\cite{distillation} with update rule specified by:
\begin{equation}
    \mathbf{x}_s = \alpha_s\hat{\mathbf{x}}_\theta(\mathbf{x}_t)  + \sigma_s\frac{\mathbf{x}_t -\alpha_t\hat{\mathbf{x}}_\theta(\mathbf{x}_t)}{\sigma_t}. 
\end{equation}

For a fair comparison, all diffusion models are trained using the same score-matching strategy and fixed number of 512 time steps during sampling.

The fast point cloud diffusion model (FPCD) follows~\cite{mikuni:point_clouds}, where a permutation equivariant estimation of the score function is obtained by the combination of a \textsc{DeepSets}~\cite{deepsets} architecture with attention layers~\cite{attention}. During the point cloud simulation, two models are also defined: one that learns the number of non-empty cells, conditioned on the initial energy of the incoming particle, and one model that learns the score function of the normalized point cloud, also conditioned on the energy of the particle to be simulated and the number of hits to be generated. This model is trained on Dataset 1, described in Sec.~\ref{sec:dataset}.

The model trained on the image dataset (\textsc{CaloScore}) is adapted from \cite{mikuni:caloscore} with a few modifications. Compared to the original implementation, the calorimeter simulation task is now broken down in two diffusion models: one that learns only the energy deposits in each layer of the calorimeter, conditioned on the initial energy of the particle to be simulated, and one model that learns to generate normalized voxels per layer, conditioned on the energy deposition in each layer and the initial energy of the particle to be simulated. Additionally, the original \textsc{U-Net}~\cite{unet} model is combined with attention layers. These changes increase the model expressiveness and the generation fidelity. This model is trained on Dataset 2, described in Sec.~\ref{sec:dataset}

\section{Detector and Data Descriptions}
\label{sec:data}

\subsection{Calorimeter Simulation}
The \textsc{DD4HEP} framework~\cite{Frank:2014zya} is used to run \textsc{Geant} simulations of a high-granularity iron-scintillator calorimeter (based on the CALICE-style design~\cite{CALICE:2022uwn}), which has dimensions similar to those of the forward hadronic calorimeter in the future ePIC detector (LFHCAL~\cite{Bock:2022lwp}) at the EIC. Specifically, the sampling structure comprises 0.3 cm scintillator tiles sandwiched between 2.0 cm thick steel plates. It consists of a total of 55 layers. The transverse area of the scintillator is set to 10 cm$\times$10 cm, somewhat larger than in Ref.~\cite{Bock:2022lwp}. It adopts a non-projective geometry with tower elements arranged in parallel to the $z$ axis and has its front face at $z$=$3.8$ m.

1.7 million events of single $\pi^+$ particles incident on the center of the calorimeter are simulated. The incident momentum, $P_\mathrm{Gen.}$, was generated uniformly in $\log_{10}$ space in the range $1.0 < P_\mathrm{Gen.} < 125$ GeV/$c$. In order to hit the center of the calorimeter, the pions were generated with a polar angle of $\theta_\mathrm{Gen.} = 17^\circ$. Because the detector is symmetric about $\phi$, the particles are generated in the range $0^\circ < \phi_\mathrm{Gen.} < 360^\circ$. 
An energy threshold corresponding to 0.3 MeV are used to select hits for further analysis.

\subsection{Datasets}
\label{sec:dataset}
Dataset 1 is the point cloud representation of the \textsc{Geant} showers, while Dataset 2 represents the same showers using the image representation. Both Dataset 1 and Dataset 2 used in training share the same parent \textsc{Geant} simulation, such that the fast point cloud diffusion model and the image model are trained on different representations of the same set of calorimeter showers.

Dataset 1 is created by taking the \textsc{Geant} simulation and converting it to a format based on JetNet data \cite{anonymous_2021_4834876}, that stores information on jets and their constituents in a zero-suppressed point cloud representation. The \textsc{Geant} data is stored in files containing two datasets, \textit{clusters} and \textit{cells}. The cluster dataset contains the $P_\mathrm{Gen}$ of the incident pion, as well as the number of hits in the calorimeter. The cell data is  comprised of a constant number of 200 cells per event. Empty cells, or cells with deposited energy below the threshold are masked, with all values set to 0.0, and ignored during training.

The $x$, $y$, and $z$ distributions of the \textsc{Geant} simulation are initially discrete, resulting from the digitization step of the simulation, with values equal to the centers of the cells in each dimension. The point cloud model struggles to learn extremely sharp features, as the score function is not well-defined for discrete inputs. To circumvent this, a uniform smearing within a cell-width is applied to the cells along each dimension to obtain continuous distributions for the final point cloud dataset. This maintains the same distributions at histogram-level when binning according to the cell-width, but yields a point cloud dataset with smooth $x$, $y$, and $z$ distributions. Without this smearing, the distributions in $x$, $y$, and $z$ resemble a series of delta functions that the point cloud model struggles. The point cloud model is trained on this smeared point cloud representation of the \textsc{Geant} simulation. 

Dataset 2 is created by converting the point cloud dataset into an image format. Images at the original granularity would would be too large for the generative model. The calorimeter cells were therefore clustered into groups of 5 along each axis of the detector to create voxels, where $5\times5\times5$ cells = 1 voxel. Energy in each of the cells making up the voxel were summed and assignd to the final voxel's total energy. The final image format consists of $11\time 11\times 11$ voxels. A hit in the voxelized dataset, and referenced in Section \ref{sec:results}, is defined as any voxel with energy deposition above threshold. 

For the final comparison, generated samples from the point cloud model are voxelized using the same method for Dataset 2. All comparisons are in this image format, at the same resolution of $11 \times 11 \times 11$ voxels per image.

Images representing the full resolution of the calorimeter with $55\times55\times55$ voxels were not used, as this would result in unmanageably large datasets (see Table \ref{tab:main_comparison}), and would represent the largest calorimeter image training ever done. The point cloud model was trained on the full resolution because point clouds naturally represent the calorimeter at full granularity. Training the point cloud model on this more natural representation is in line with the goal of this work to investigate advantages/disadvantages of two representations of the calorimeter data. It is also for this reason that the generated point cloud distributions are shown separately, while the direct comparisons between models are done in the image representation.  Investigating possible advantages of a point-cloud model trained directly on the voxelized dataset is left to future work.

\section{Results}
\label{sec:results}
\begin{figure*}[] 
        \subfloat[]{\includegraphics[width = \textwidth/2]{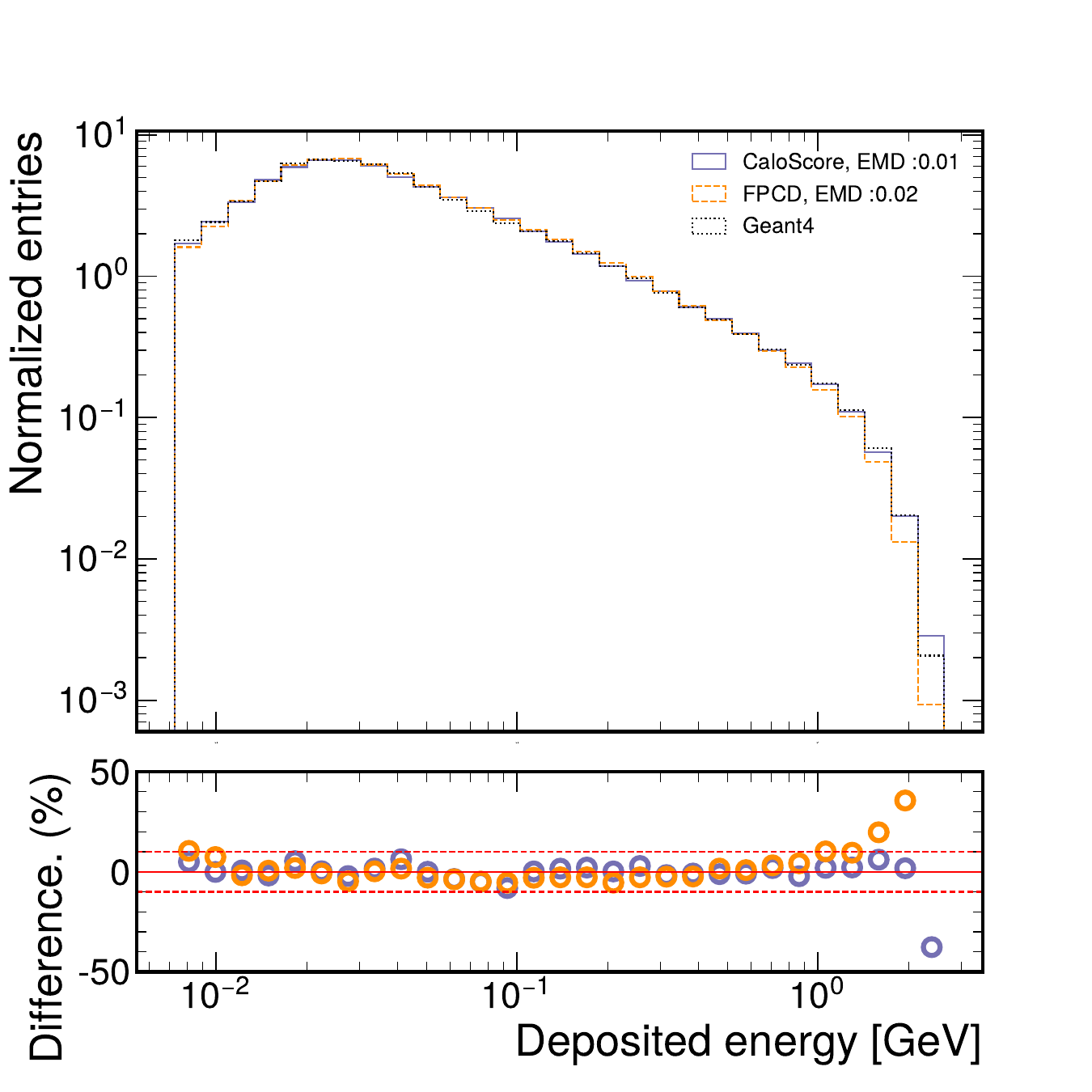}}
        \subfloat[]{\includegraphics[width = \textwidth/2]{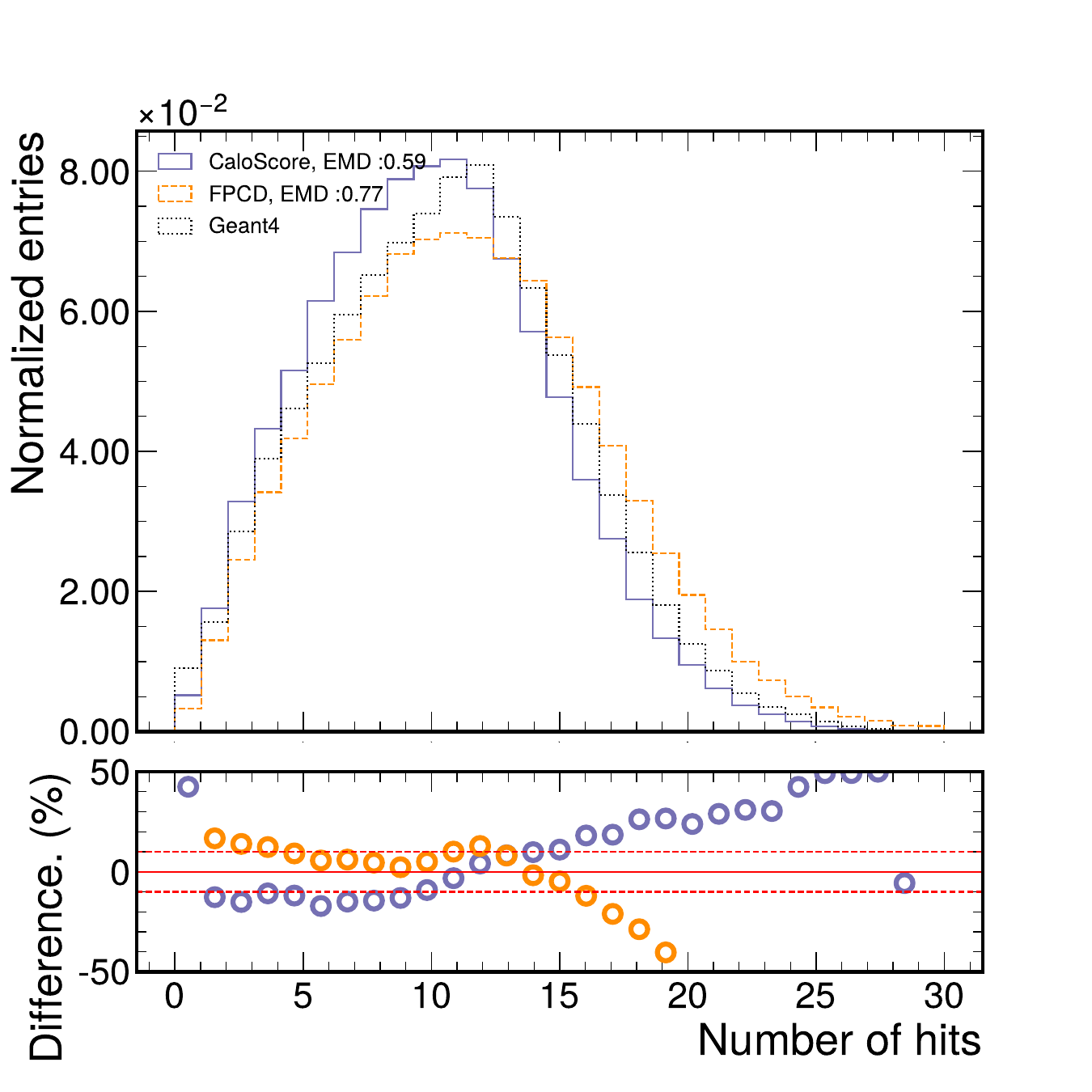}}
    \caption{Sum of all voxel energies of the image representation of FPCD model, shown in orange,  and the image based model, shown in grey-blue. The parent \textsc{Geant} distributions are shown as a dotted black line in the top panels. The dashed red lines in the bottom panel of each figure represent the 10\% deviation interval of the generated samples from the original \textsc{Geant} simulation. The earth mover’s distance (EMD) between each distribution and the \textsc{Geant} distribution is also shown.}
    \label{fig:total_e_and_hits}
\end{figure*}

\begin{figure*}[] 
        \subfloat[]{\includegraphics[width = \textwidth/2]{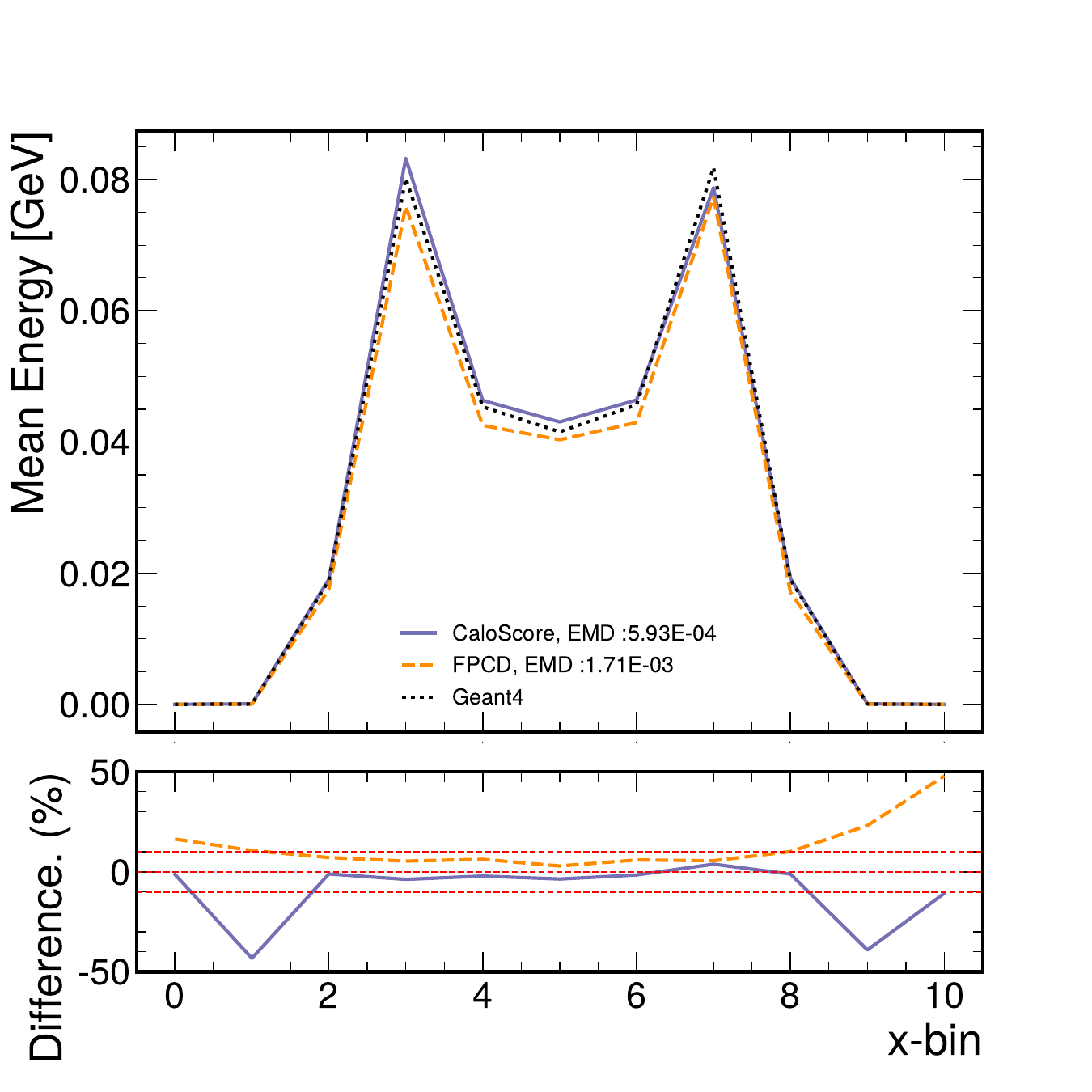}}
         \subfloat[]{\includegraphics[width = \textwidth/2]{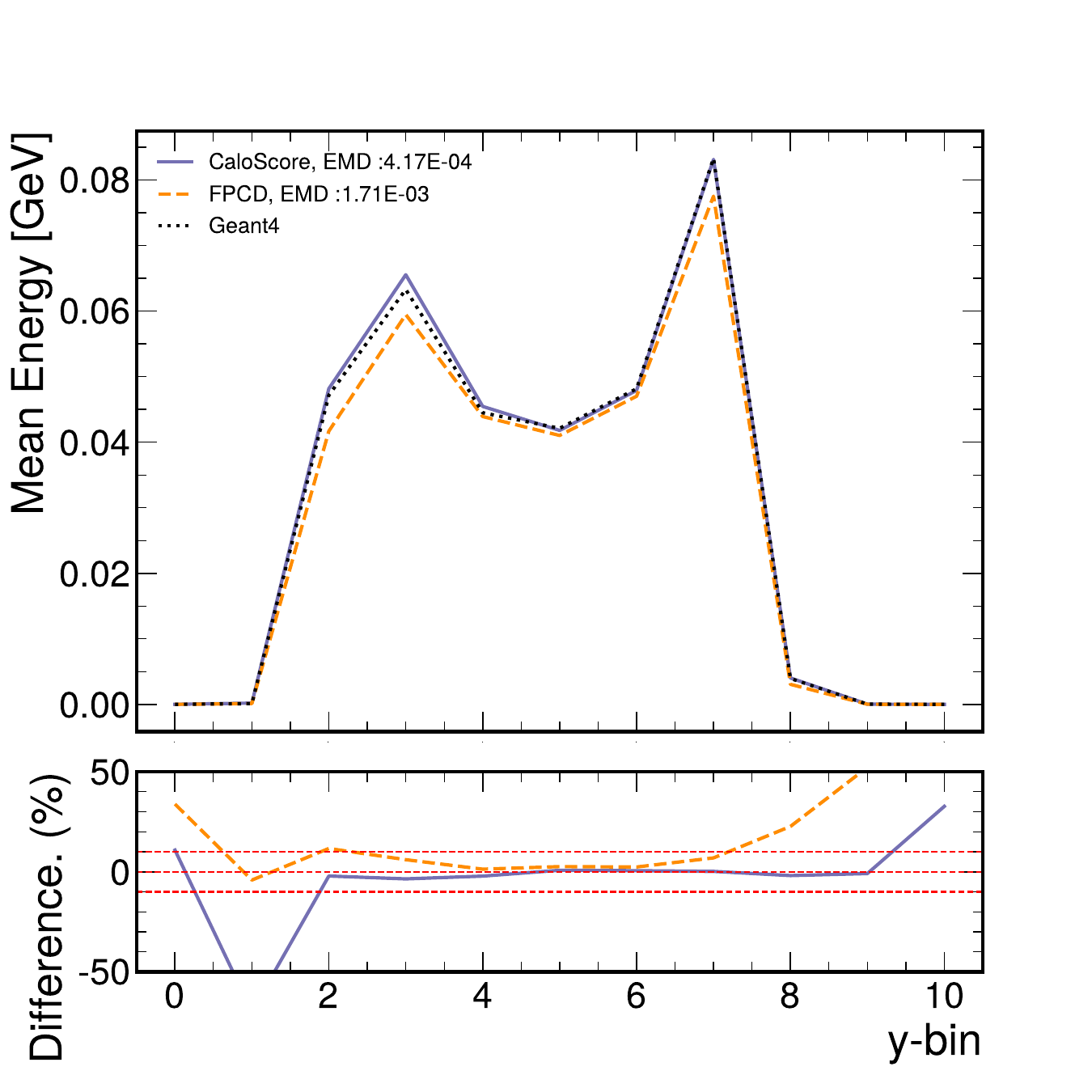}}
    \caption{Comparison of the average deposited energies in the $x$ (left), $y$ (center), and $z$ (right) coordinates. The dashed red lines in the bottom panel of each figure represent the 10\% deviation interval of the generated samples from the original \textsc{Geant} simulation.}
    \label{fig:image_xy}
\end{figure*}

\begin{figure}[h] 
        {\includegraphics[width = \textwidth/2]{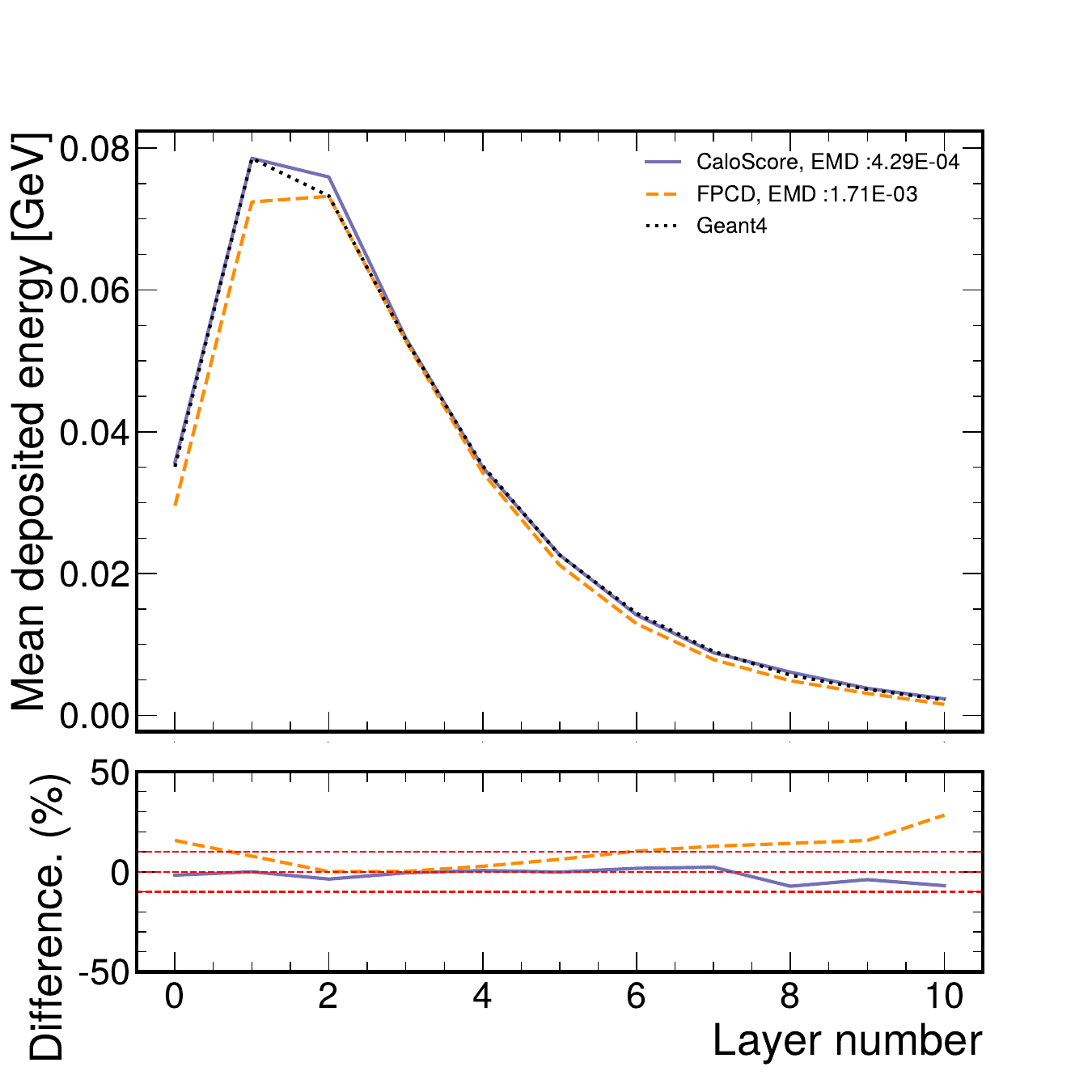}}
    \caption{Comparison of the average and z coordinate. The dashed red lines in the bottom panel of each figure represent the 10\% deviation interval of the generated samples from the original \textsc{Geant} simulation.}
    \label{fig:image_z}
\end{figure}

All generated samples along with \textsc{Geant} are converted to the same image format at the same resolution of $11\times 11\times 11$ voxels per event for fair comparison. A variety of distributions are used to evaluate the quality of the generated images. After comparing calorimeter images generated by both models, the point cloud representation of \textsc{Geant} is compared to the generated samples of the point-cloud model to provide additional insight to the previous image-based comparison. For all comparisons, the Earth mover's distance (EMD) \cite{EMD}, also known as the 1-Wasserstein distance \cite{EMD-W1}, between generated distributions and \textsc{Geant)} distributions is calculated.
The EMD score a distance-like measure of the dissimilarity between two distributions. It roughly represents the minimum amount of work needed to transform one distribution to another. While this is not the only possible metric, it is a standard and widely-used statistic that was also the main distance deployed in \cite{mikuni:caloscore}, where an image based model was compared to a Wasserstein-GAN. All EMD scores in Figures~\ref{fig:total_e_and_hits}, \ref{fig:image_xy} and \ref{fig:image_z} are calculated on final voxelized distributions 

\begin{figure*}[] 
    \subfloat[]{\includegraphics[width = 2.2in]{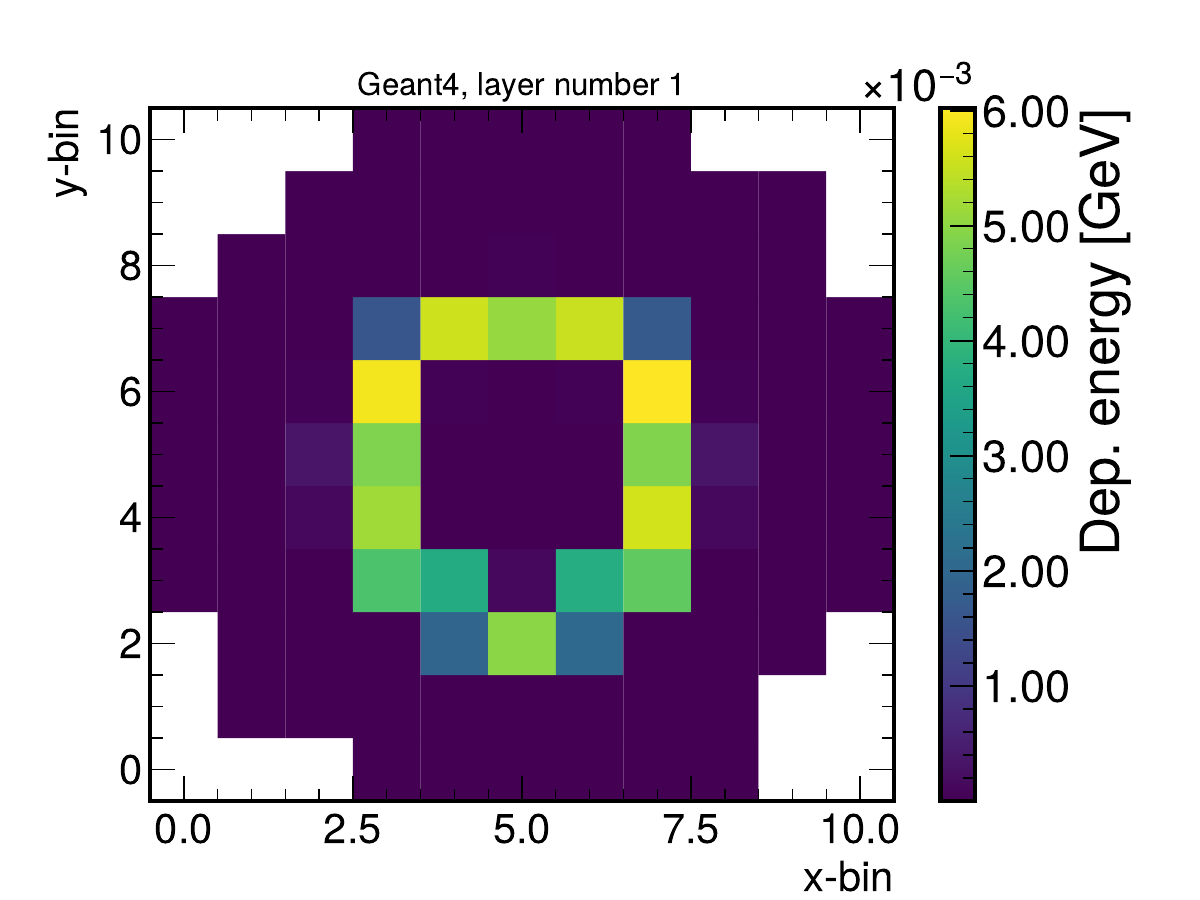}}
    \subfloat[]{\includegraphics[width = 2.2in]{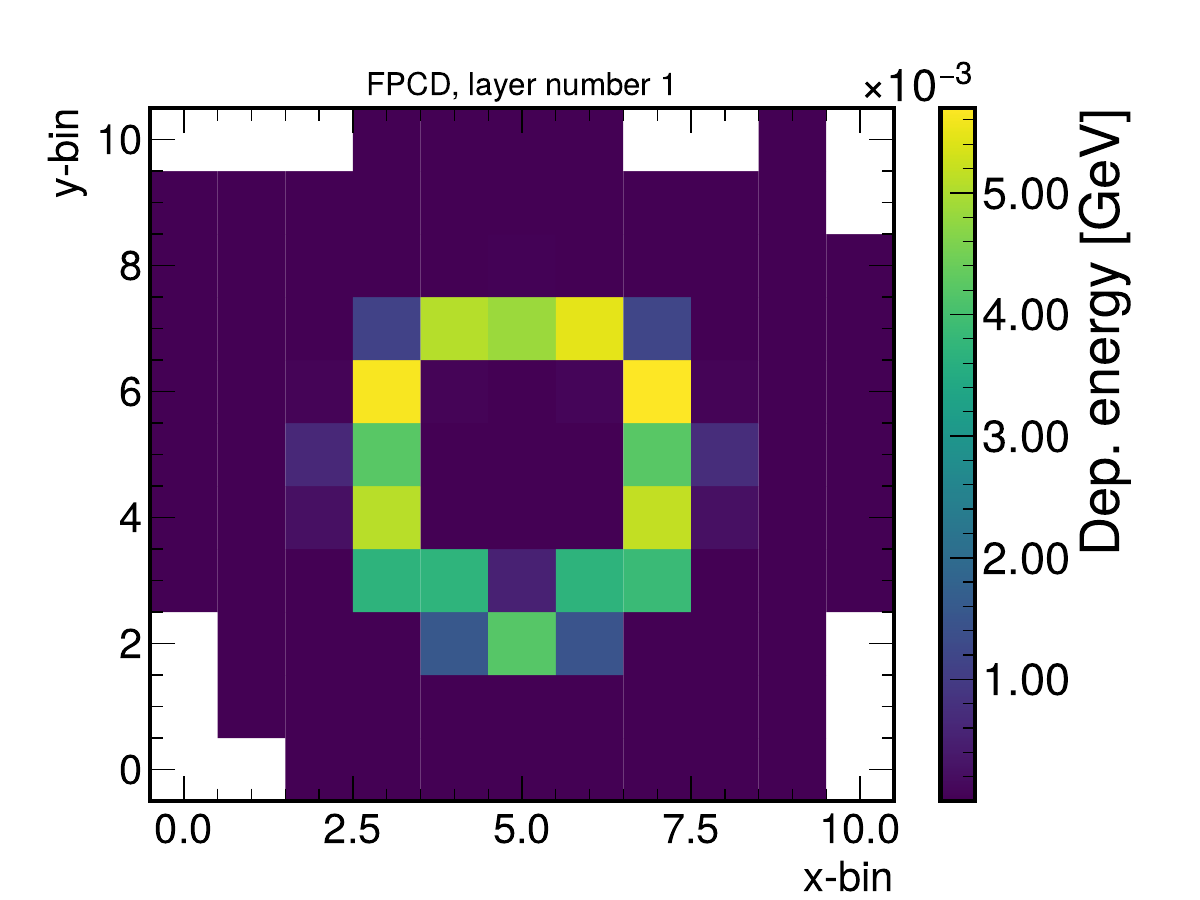}}
    \subfloat[]{\includegraphics[width = 2.2in]{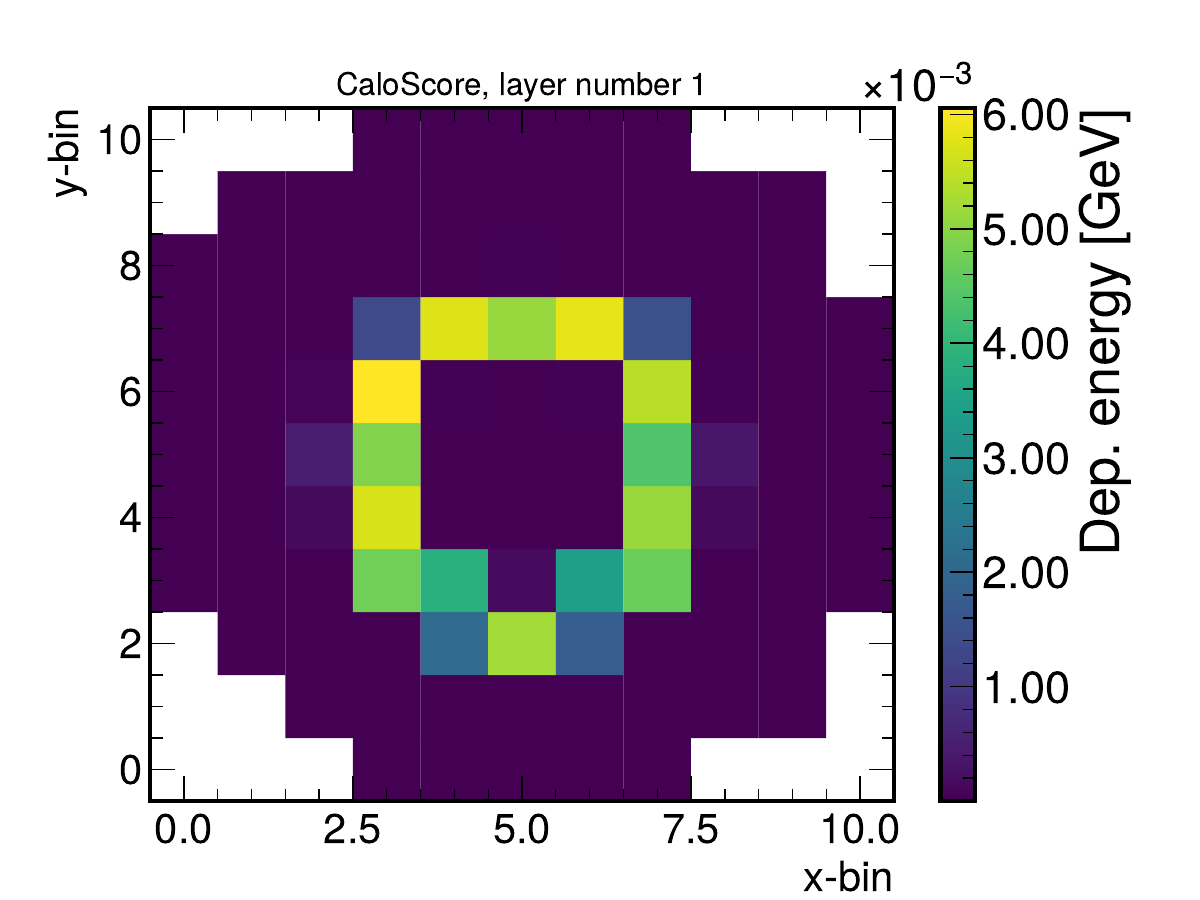}}\\
    \subfloat[]{\includegraphics[width = 2.2in]{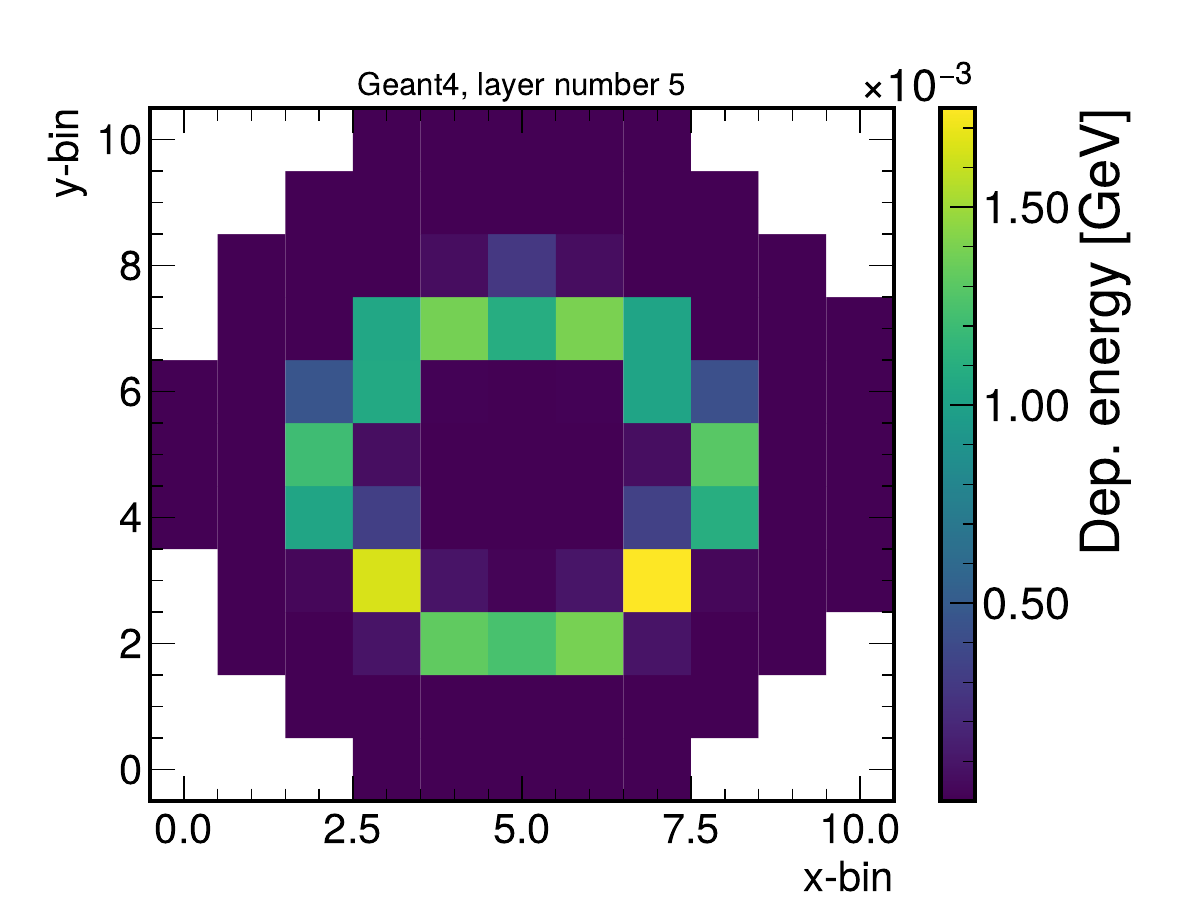}}
    \subfloat[]{\includegraphics[width = 2.2in]{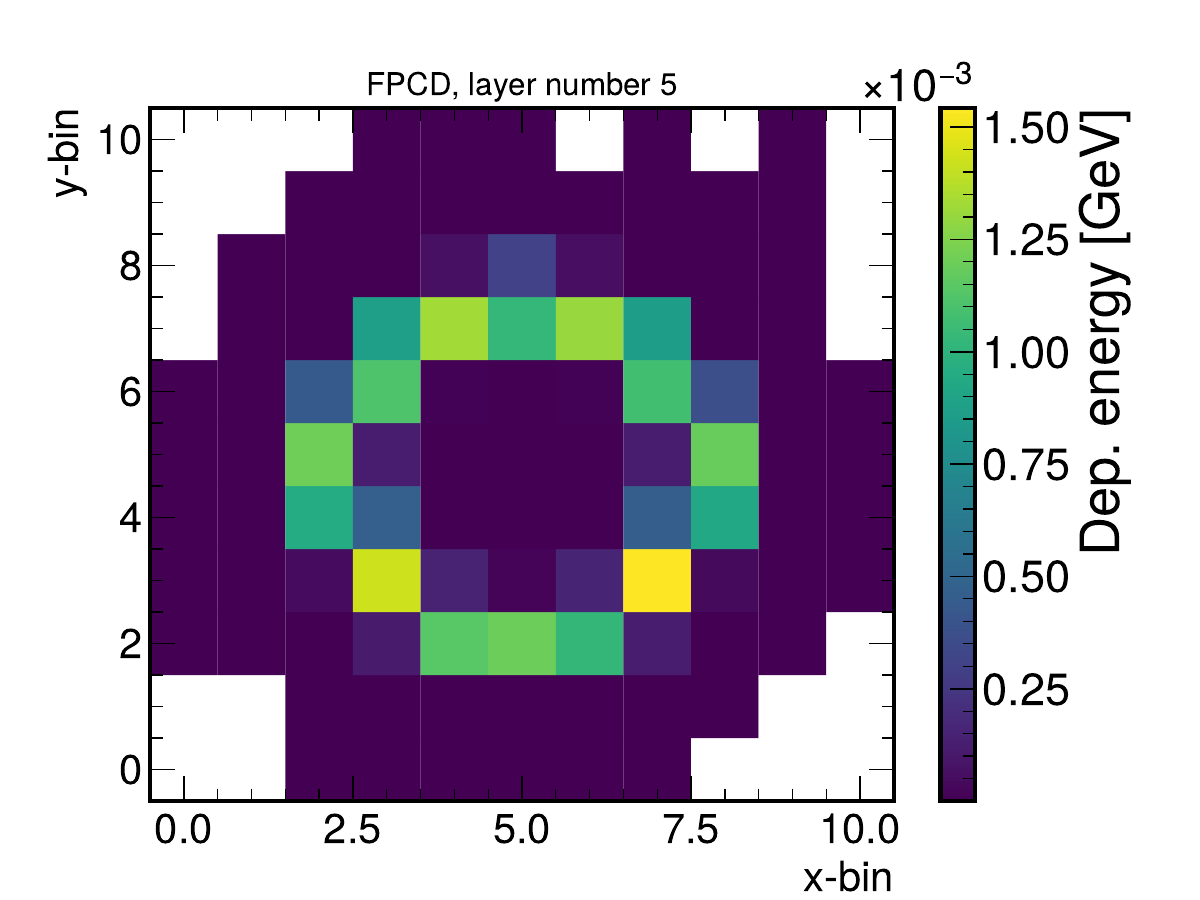}}
    \subfloat[]{\includegraphics[width = 2.2in]{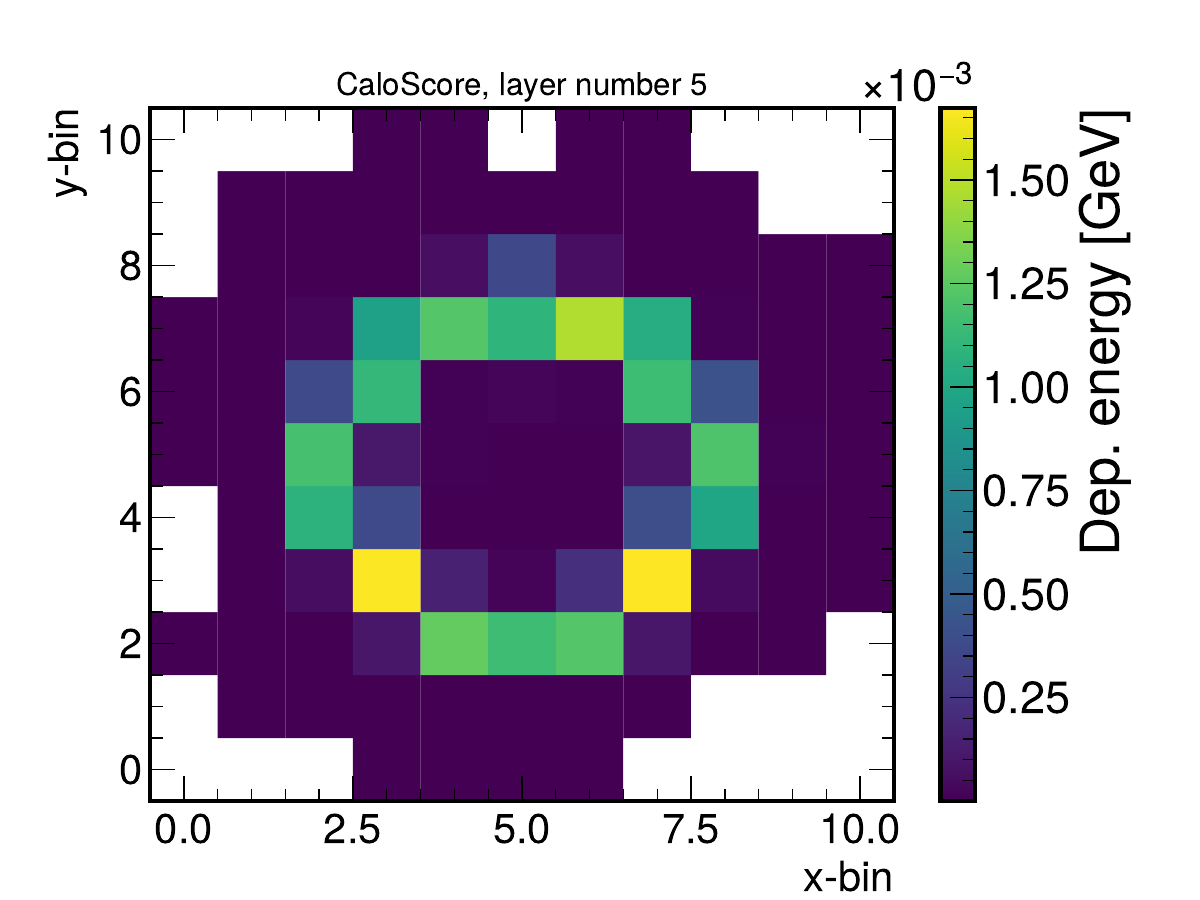}}\\
    \subfloat[]{\includegraphics[width = 2.2in]{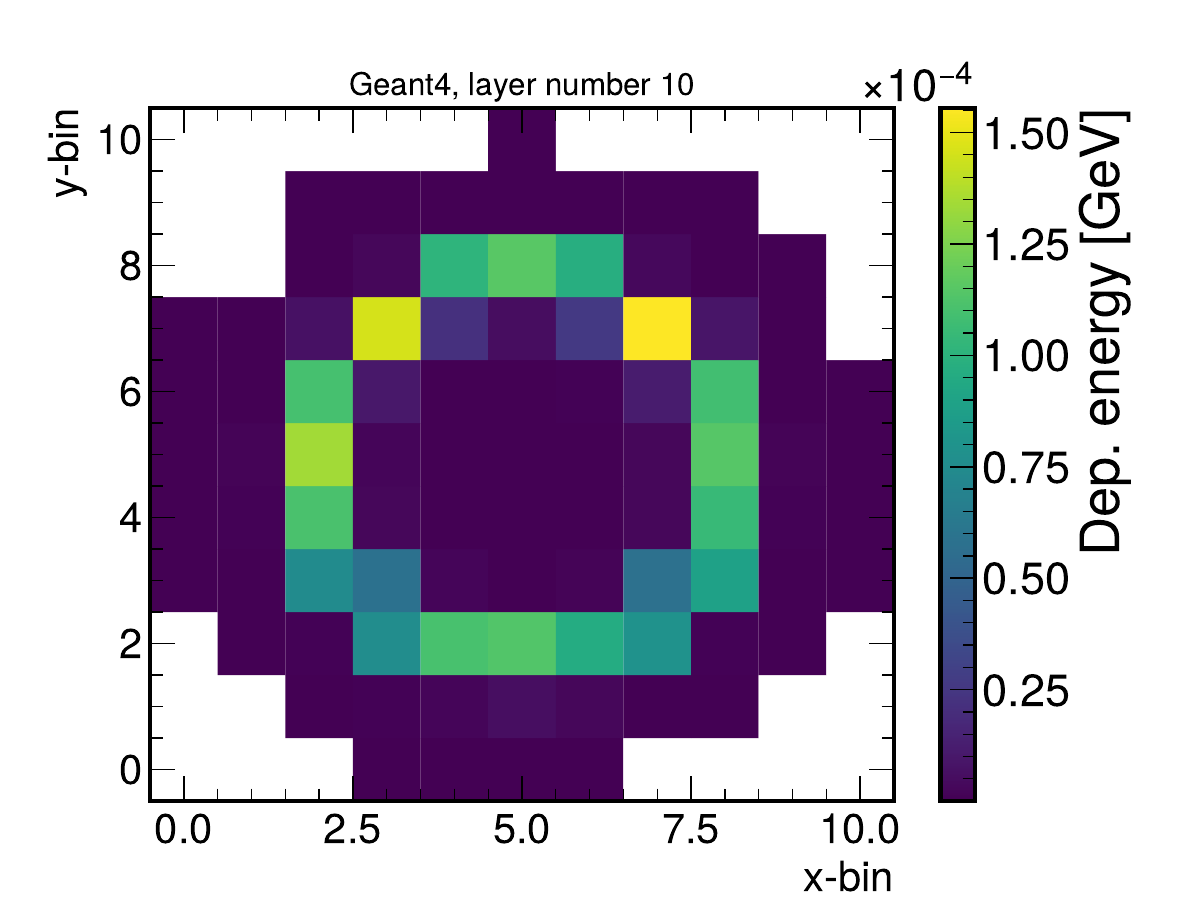}}
    \subfloat[]{\includegraphics[width = 2.2in]{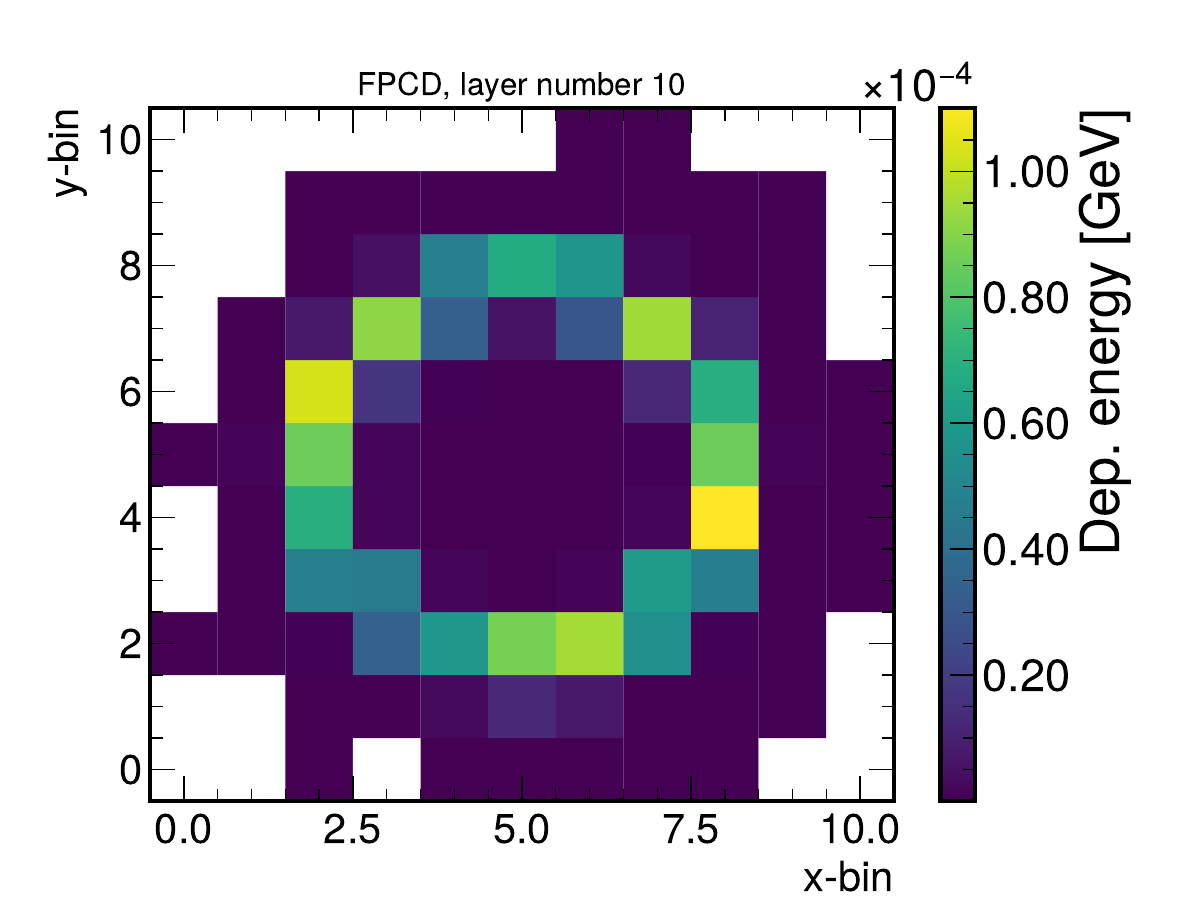}}
    \subfloat[]{\includegraphics[width = 2.2in]{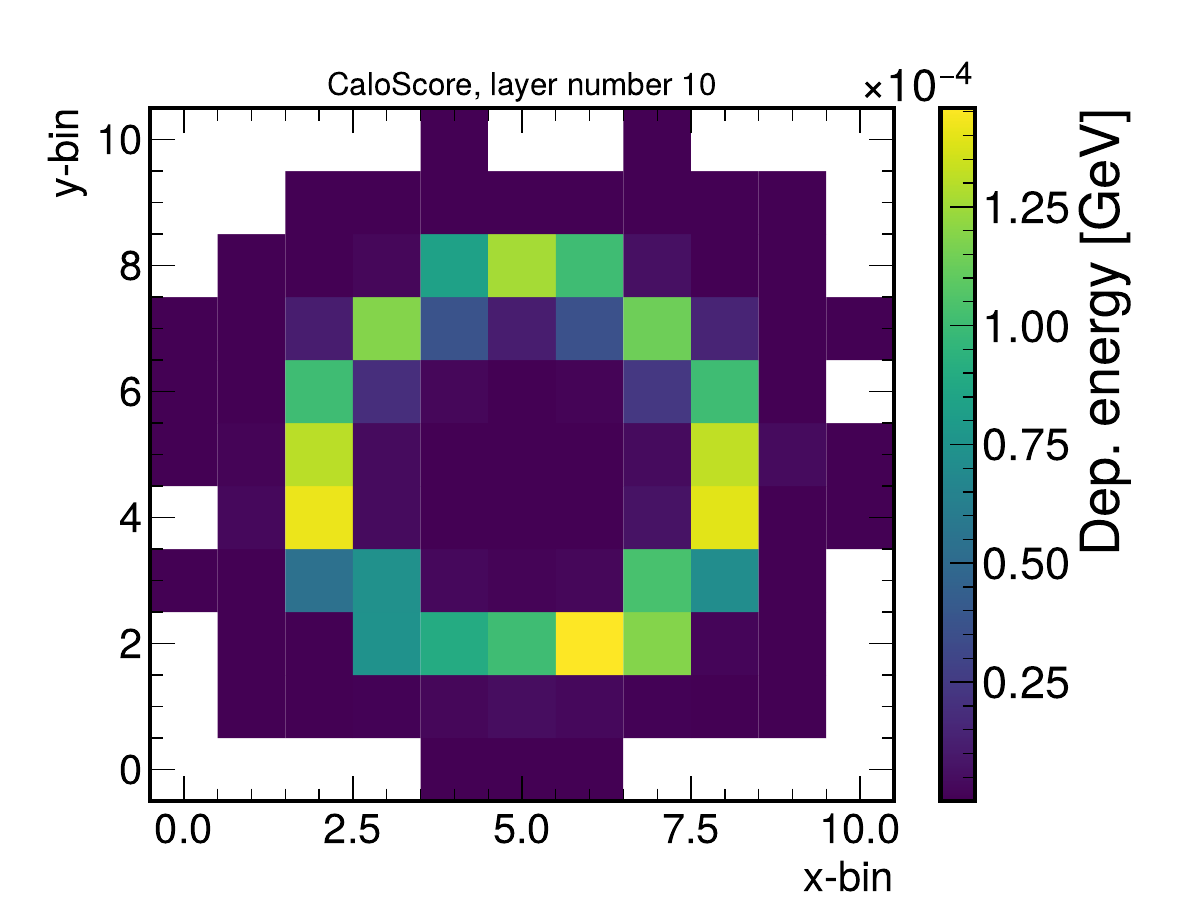}}\\
    \caption{The 2-dimensional distribution of the mean deposited energy in the 1st, 5th, and 10th voxelized layer of the calorimeter. The first column is the original Geant simulation. The second column is the fast point-cloud based diffusion model (FPCD), and the 3rd column is the image-based model (\textsc{CaloScore}).}
    \label{fig:2d_images}
\end{figure*}

Figure~\ref{fig:2d_images} shows a qualitative assessment of the generative models using the 2-dimensional distribution of the average energy deposition in three layers. All voxels with an expected energy deposition above 0 are populated in both the image and point cloud based models, with very few additional hits. The calorimeter shower will have diverse shapes, as well as different overall distribution of voxels due to the variation of $\phi_\mathrm{Gen.}$.  The qualitative similarities in each image in Fig~\ref{fig:2d_images} indicate that models reproduce the various showers from the training dataset well. Each image contains a ring due to $\theta_\mathrm{Gen.}$ being fixed while varying $\phi_\mathrm{Gen.}$.

Table \ref{tab:main_comparison} shows the model size, size of each dataset, and time to generate 100k calorimeter showers. The disk size and sample time under the point cloud model are for showers in the point cloud representation. The AUC is obtained from a classifier trained to distinguish the samples of both models only in the voxelized image format. Both models have very good AUC, reasonably close to 0.5, with the image model having the lower AUC. The point cloud model is smaller by a factor of 4 compared to the image based model, and samples events 3 times faster. Lastly, the point cloud dataset requires over 100 times less disk space than the image format at full granularity.

\begin{table*}[]
\begin{tabular}{@{}l|l|l|l|l|l|l@{}}
\toprule
 & Model & \# Parameters & Disk Size (Full) & Sample Time& AUC \\ \midrule
 & Image        & 2,572,161 & 1016MB (62GB) & 8036.19s & 0.673 &\\
 & Point Cloud  & 620,678   & 509 MB        &   2631.41s      & 0.726 &\\ \bottomrule
\end{tabular}
    \caption{Comparison of model size, size of data representation on disk, generation time, and AUC of the same classifier trained to distinguish between the model and the original \textsc{Geant} showers. All comparisons are done for 100k calorimeter showers. The all results in the image row were obtained with the scaled down, $11\times11\times11$ voxel images, however the disk size of the image dataset at full granularity is shown in parenthesis.}
    \label{tab:main_comparison}
\end{table*}

Figure~\ref{fig:total_e_and_hits} compares the total energy deposited in the calorimeter and total number of calorimeter hits, where a hit is defined as any voxel with energy above threshold. The EMD is also calculated between \textsc{Geant} and the different generative models.

Both the image-based diffusion model and the point-cloud based diffusion model are in good agreement with \textsc{Geant} at small deposited energies, deviating no more than 10\%. At the highest deposited energies, however, both diffusion models begin to fall away from \textsc{Geant}, with the point-cloud model generating less energy, and the image based model generating slightly more energy than \textsc{Geant}. These trends begin at about 10 GeV, with the point-cloud model deviating slightly earlier. The point-cloud model also shows a slightly higher EMD score than the image based model. 
The region where the deviations are largest, past 20~GeV of deposited energy are rare, and statistical fluctuations begin to dominate the \textsc{Geant} distributions. 

The number of hits shows a similar trend, though with larger deviations. At a small number of hits, both show good agreement with \textsc{Geant}, with deviations slightly above 10\%. At 15 or more hits, both models begin to deviate well past 10\%, with the point cloud model oversampling the number of hits, and the image based model generating less hits than \textsc{Geant}.

Figure~\ref{fig:image_xy} and~\ref{fig:image_z} shows the average deposited energy  $x$, $y$, and $z$-coordinates. Both models struggle in the first and last layers in $x$ and $y$ coordinates, but show good agreement in the middle layers. While the image-based model shows larger deviations in the first and last layers of the calorimeter compared to the point-cloud model, it has an overall lower EMD in both distributions. The two-pronged feature of these distributions is a result of generating the pions at a fix polar angle and varying $\phi$.  It should be noted that there are little to no hits in the first and last $x$ and $y$ layers of the calorimeter, so even a very small deviation from \textsc{Geant} will result in a large deviation percentage (bottom panels of Fig.~\ref{fig:image_xy} and~\ref{fig:image_z}).  Similarly, as there are fewer hits towards the back of the detector, deviations increase slightly for the very last layers. However, The $z$-distributions show both models in very good agreement with the original \textsc{Geant} predictions, a possible effect of the $z$-distribution of hits being less dependant on the generated $\theta$ and $\phi$ ranges. 

All three distributions show the point cloud samples are systematically lower than the original \textsc{Geant} distributions. This indicates the point cloud model would benefit from learning the energy per layer directly, as is done in the image model described Sec.~\ref{sec:DLmodels}. This difference likely explains why this small bias is observed in the point cloud model, but not in the image model, and is an avenue for improving the point cloud.

Following \cite{caloflow1}, a classifier was trained to distinguish between generated showers and \textsc{Geant} showers. The classifier is comprised of two fully connected layers of size 256 using the RELU activation function. The classifier is trained only on vectors of voxelized images of each dataset. The area under the receiver-operator curve (AUC) for the image model was 0.673.  The AUC for the point-cloud model was 0.726.  Generally, being closer to 0.5, where the classifier is maximally confused, is the target. However the AUC obtained by both models is very promising, as having an AUC even slightly below 1.0 is non-trivial.

\begin{figure}[h]
         {\includegraphics[width = 3.7in]{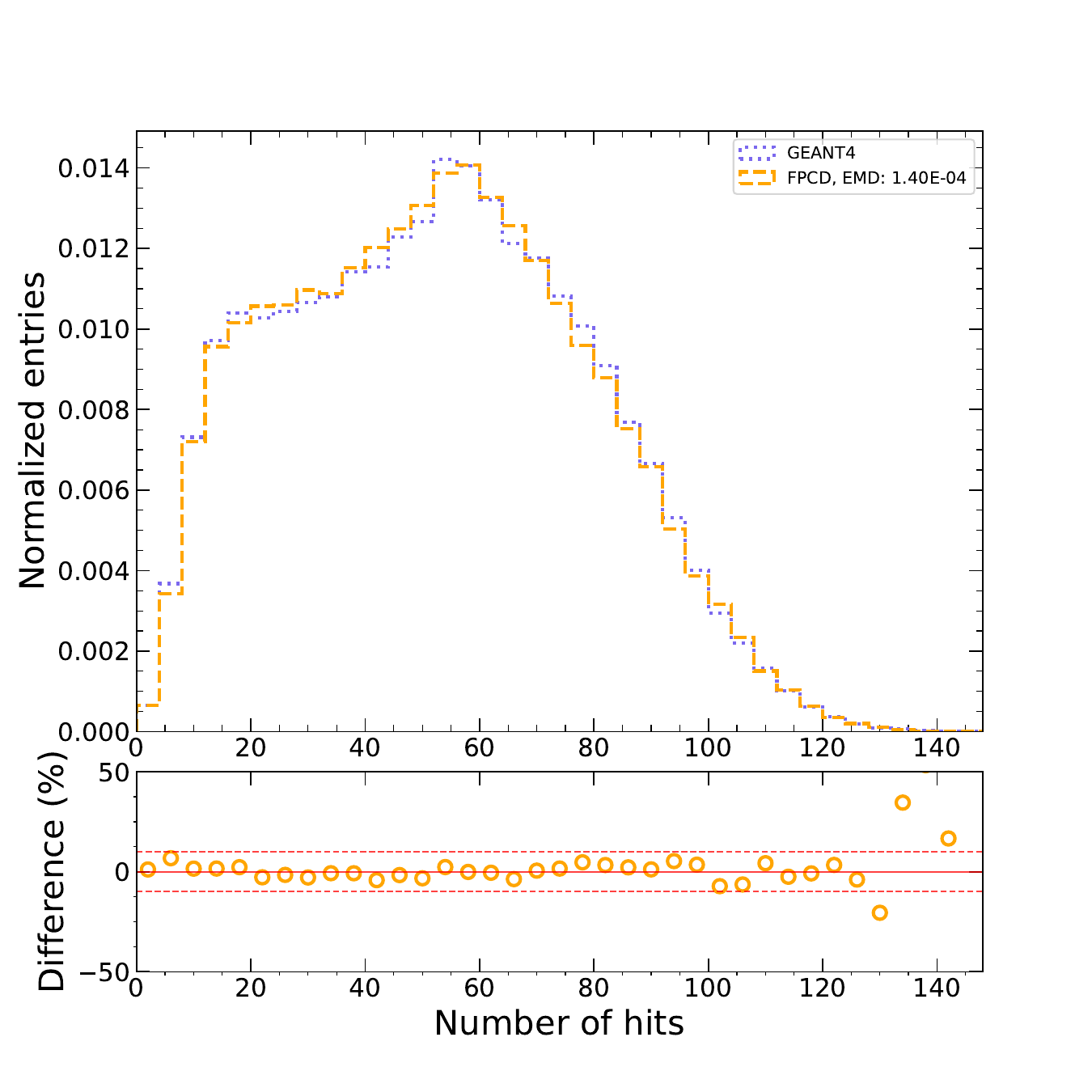}}
    \caption{The total number of hits in the point cloud representation of calorimeter showers, at full granularity. The dashed red lines in the bottom panel of each figure represent the 10\% deviation interval of the generated samples from the original Geant simulation.}
    \label{fig:pointcloud_nhits}
\end{figure}

A key advantage of the point cloud model is that the distributions at the sub-voxel level can be shown. The point cloud model already simulates the data at the original granularity of the calorimeter, and voxelization is only necessary for the image representation. The original output of the point cloud model is compared to the continuous (or smeared) \textsc{Geant} distributions.
Figure \ref{fig:pointcloud_nhits} shows the number of hits in the point cloud representation of the calorimeter showers. In the point-cloud representation, a hit is defined as any \emph{cell} that has a energy deposited above threshold. 

The point-cloud model reproduces the total number of cell hits well, much better than the voxel hit distribution, shown in Fig.~\ref{fig:total_e_and_hits}. This may indicate that while the point cloud model is overall similar to \textsc{Geant} in both representations, small deviations in point cloud distributions can be summed into larger deviations during the voxelization process, where 125 individual cells are combined into a single voxel. However, there is a large symmetry group under which mismodelings in the bigger space may not affect the modeling in the coarser space, so further investigation is needed. However, the very good agreement with \textsc{Geant} in the number of cell hits and degrading agreement in the number of voxel hits indicates that the first diffusion model of the point cloud model architecture is performing well, while the second model, responsible for sampling the cell distributions, would likely benefit from additional tuning.

\begin{figure*}[h] 
        \subfloat[]{\includegraphics[width = \textwidth/2]  {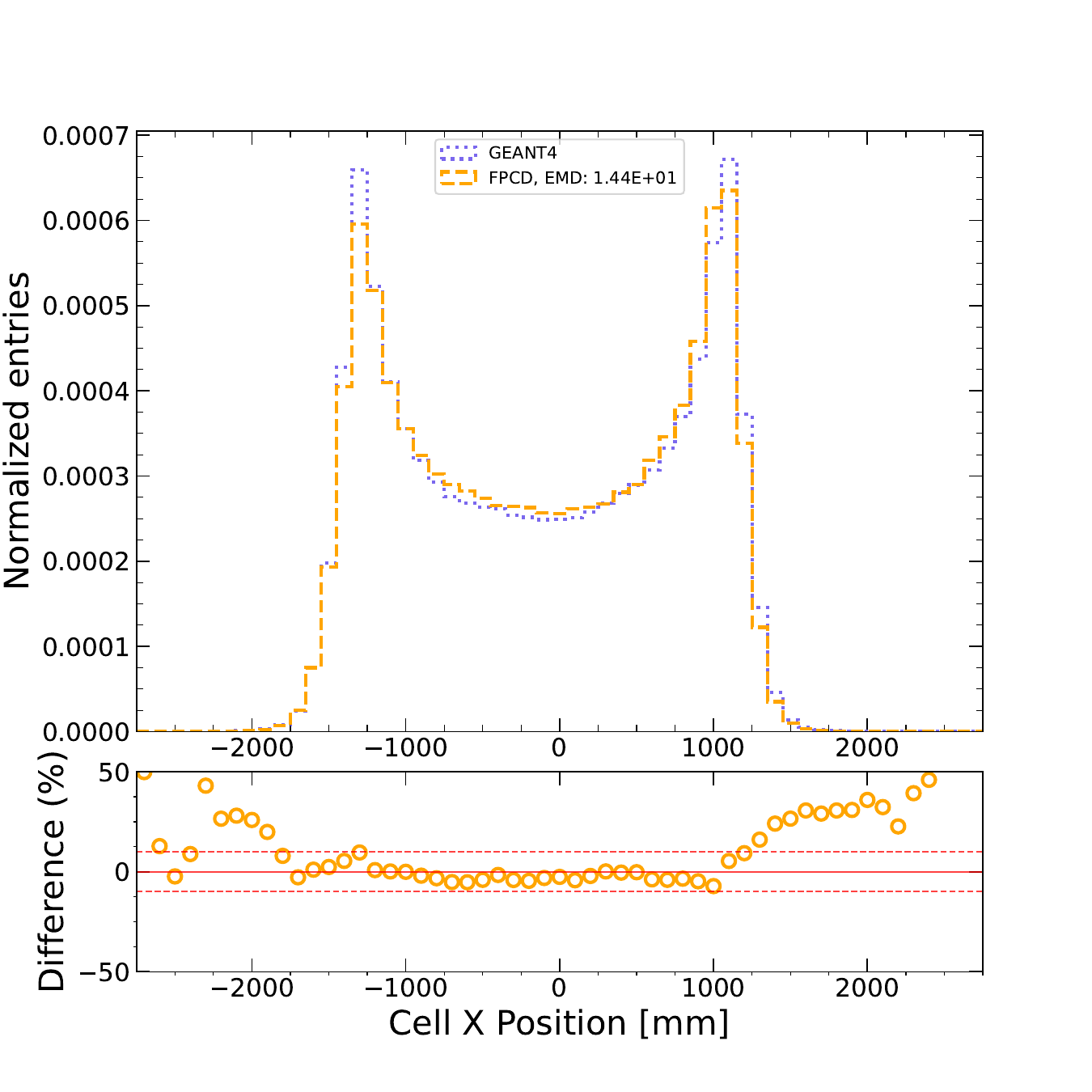}}
         \subfloat[]{\includegraphics[width = \textwidth/2]{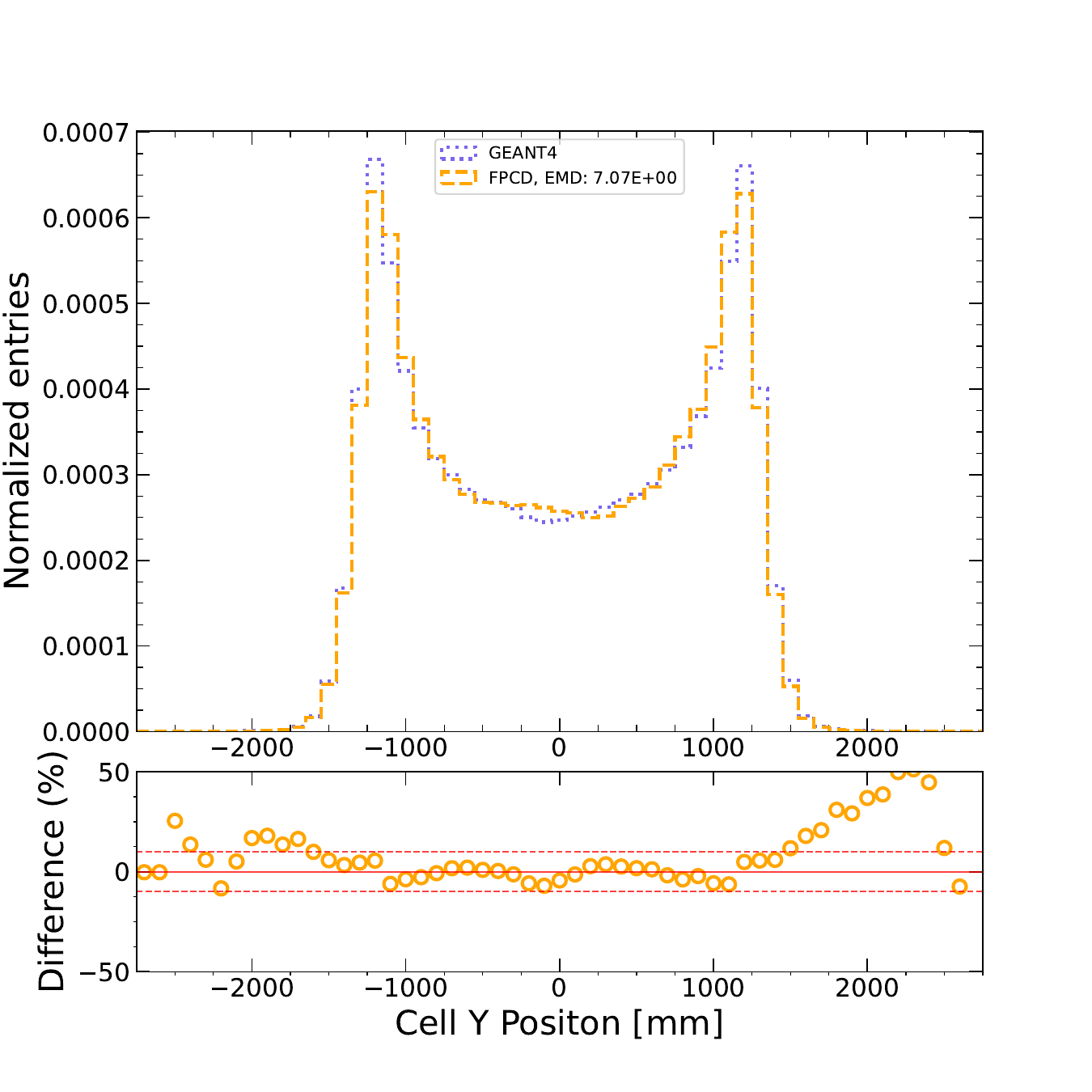}}\\
         \subfloat[]{\includegraphics[width = \textwidth/2]{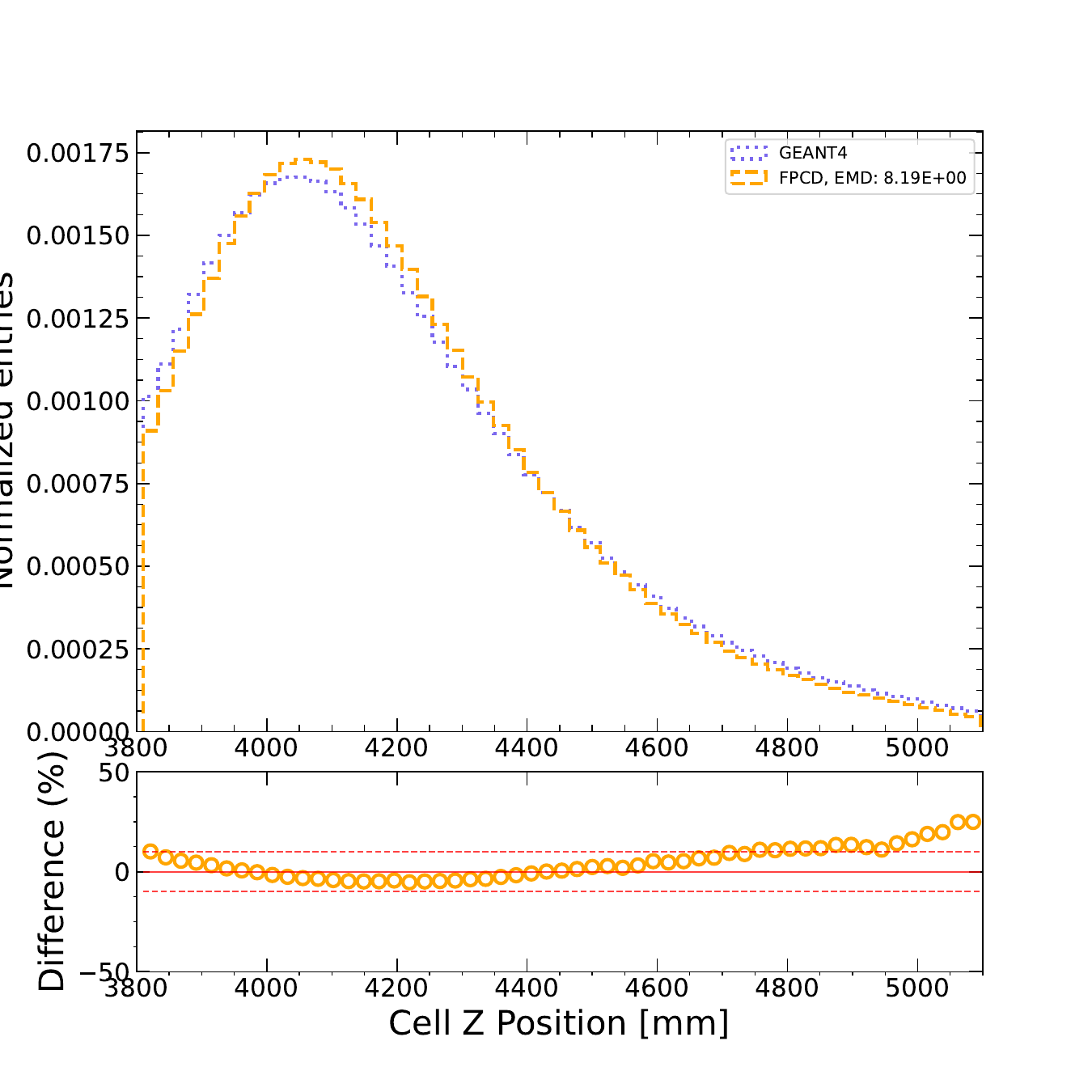}}
         \subfloat[]{\includegraphics[width = \textwidth/2]{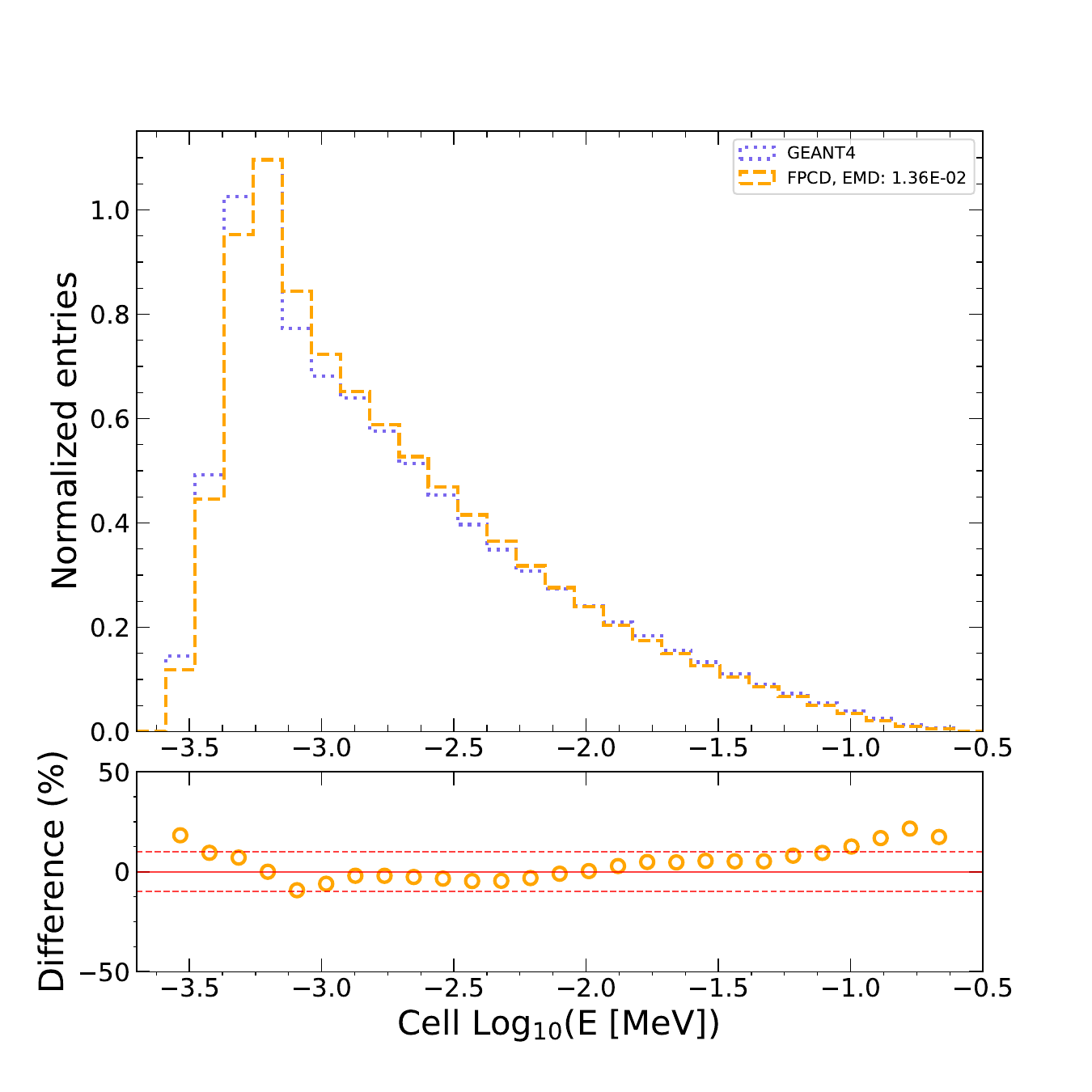}}
    \caption{Comparison of the average cell $x$ (top left), $y$ (top right), $z$ (bottom left) and $\log_{10}$E (bottom right) distributions in the point cloud datasets. Each distribution is binned according to the cell-width to show the full granularity of the detector. The dashed red lines in the bottom panel of each figure represent the 10\% deviation interval of the generated samples from the original \textsc{Geant} simulation.}
    \label{fig:point_cloudxyz}
\end{figure*}

Similar conclusions can be derived from Fig.~\ref{fig:point_cloudxyz}, show the generated point samples at the full detector granularity and in good agreement with \textsc{Geant}. Fig.~\ref{fig:point_cloudxyz} shows the average $x$, $y$, and $z$ coordinate distributions, as well as the cell $\log_{10}$E distribution in the point representation. Again, there are larger relative deviations in the first and last layers in $x$, $y$, and $z$, coordinates where there are very few hits, just as in the image representation. However, there is very good agreement with the \textsc{Geant} simulation in layers containing a reasonable number of hits.

\section{Conclusion and Outlook}

In this paper, we make the first direct comparison between two score based generative models using either images or point clouds as representations of the same training data. We use \textsc{Geant} calorimeter simulations of a high-granularity hadronic calorimeter. Both models perform well for most distributions, with very similar AUCs, but the image-based diffusion model invariably has a lower EMD in each comparison to \textsc{Geant}. 

Overall, the performance of the point-cloud diffusion model is very close to the image model. This is despite the point cloud model being disadvantaged in this work in a few important ways. 

First, the calorimeter showers from the FPCD model are closest to \textsc{Geant} in the point cloud representation at the full calorimeter granularity, as shown in Fig.~\ref{fig:pointcloud_nhits} and \ref{fig:point_cloudxyz}. But it is later voxelized for comparison. This may compound mismodeling during the voxelization, however further investigation is needed. 

Second, the point cloud model is adapted from a model architecture initially designed for jet data from the JetNet datasets. While the high-level structure of the datasets are very similar, the data itself are quite different. For example, the first diffusion model making up the point cloud model was initially much larger, as predicting the jet multiplicity is in general a more difficult problem than the number of non-empty cells in a calorimeter shower. Reducing the size of the first diffusion model of the point cloud model architecture had no impact on performance while speeding up training. The second diffusion model making up the point cloud model architecture that is responsible for sampling the cell $x$, $y$, $z$, and $E$ was directly adapted from \cite{mikuni:point_clouds}. Further tuning of the point cloud  model, particularly the cell-model can likely close the small remaining gap in performance.  The image model, in contrast, is based on \textsc{CaloScore}, which was tuned specifically for calorimeter showers. 

Lastly, the image-based model uses the energy deposition in each layer in addition to the generated particle momentum to condition the second diffusion model making up its architecture. The second diffusion model making up the point cloud model is solely conditioned on the generated particle momentum. This might explain why the point cloud model has systematically lower mean energy distributions (see Fig.~\ref{fig:image_xy} and~\ref{fig:image_z}) compared to both \textsc{Geant} and the image based model.

These potential sources of improvement in the point cloud model should not detract from it's already very reasonable performance, deviating from \textsc{Geant} more 10\% only in the sparsest of layers, where the image based model also struggles. At the same time, the point cloud model offers several advantages over the image model.

First, the sheer size of the data. The point cloud data saved to \textsc{HDF5} files is a factor of 100 times smaller using the same zlib compression as the image based dataset at full granularity, with no voxelization. As calorimeters continue to increase in granularity, this difference will only increase. 

Second, information is lost during voxelization process; cell hits with the same $x$, $y$, $z$ coordinates, but different energies are summed over in the image representation. This is true even if images are produced at the full granularity of the calorimeter, where hits within the single cells are summed over. This means that voxelized datasets cannot naturally be reverted back to a point cloud representation.

Additionally, as was showed in this work, the generated point clouds can be voxelized afterwards, or converted into other representations that better fit specific use cases.

This work establishes a benchmark for future research on generative models, offering valuable insights into the challenges of modeling hadronic showers in highly granular calorimeters using image-based techniques, while also exploring the potential of point-cloud methods. The current advantages of point clouds, in combination with improvements to close the remaining performance gap described earlier, will likely make point cloud based models a clear choice for highly granular calorimeters. This work should serve as a reference for studies utilizing future calorimeters based on the CALICE design, including those intended for use in CMS at the LHC and ePIC at the EIC.

\label{sec:conclusions}

\section*{Code Availability}
The code used to produce the point cloud results shown in this document are available at \url{https://github.com/ftoralesacosta/GSGM_for_EIC_Calo}. The code for the image based model and comparisons of images is available at \url{https://github.com/ViniciusMikuni/Calo4EIC}. Example \textsc{Geant4} datasets and generated samples are available at \url{https://zenodo.org/record/8128598}.

\section*{Acknowledgments}

We acknowledge support from DOE grant award number DE-SC0022355.This research used resources from the LLNL institutional Computing Grand Challenge program and the National Energy Research Scientific Computing Center, a DOE Office of Science User Facility supported by the Office of Science of the U.S. Department of Energy under Contract No. DE-AC02-05CH11231 using NERSC award HEP- ERCAP0021099. M.A acknowledges support through DOE Contract No. DE-AC05-06OR23177 under which Jefferson Science Associates, LLC operates the Thomas Jefferson National Accelerator Facility. This work was performed under the auspices of the U.S. Department of Energy by Lawrence Livermore National Laboratory under Contract No. DE-AC52-07NA27344.

\bibliographystyle{unsrt} 
\bibliography{HEPML,biblio} 

\end{document}